\documentclass[letterpaper]{article} 

\usepackage{aaai24}  
\usepackage{booktabs}
\usepackage{times}  
\usepackage{helvet}  
\usepackage{courier}  
\usepackage{amsmath}
\usepackage{multirow} 
\usepackage[hyphens]{url}  
\usepackage{graphicx} 
\def\UrlFont{\rm}  
\usepackage{natbib}  
\usepackage{caption} 
\usepackage{xcolor} 
\usepackage{algcompatible}
\usepackage{algpseudocode}
\usepackage{amssymb}
\newcommand{\nj}[1]{\textcolor{black}{#1}}

\newcommand{\jh}[1]{\textcolor{black}{#1}}
\newcommand{\yr}[1]{\textcolor{black}{#1}}
\usepackage{adjustbox}

\usepackage{adjustbox}
\newcommand{\ie}{\textit{i}.\textit{e}., }
\newcommand{\eg}{\textit{e}.\textit{g}.\ }
\usepackage{lipsum}
\usepackage{algorithm}
\usepackage{layouts}
\usepackage{newfloat}
\usepackage{listings}
\DeclareCaptionStyle{ruled}{labelfont=normalfont,labelsep=colon,strut=off} 
\floatstyle{ruled}
\newfloat{listing}{tb}{lst}{}
\floatname{listing}{Listing}
\title{Any-Way Meta Learning}
\author{
    Junhoo Lee,
    Yearim Kim\equalcontrib,
    Hyunho Lee\equalcontrib,
    Nojun Kwak\thanks{Corresponding Author.}
}
\affiliations{

    Seoul National University\\
    \{mrjunoo, yerim1656, hhlee822, nojunk\} @ snu.ac.kr
}

\usepackage{bibentry}

\begin{document}

\maketitle

\begin{abstract}
Although meta-learning seems promising performance in the realm of rapid adaptability, it is constrained by 
fixed cardinality. When faced with tasks of varying cardinalities that were unseen during training, 
the model lacks its ability. In this paper, we address and resolve this challenge 
by harnessing `label equivalence' emerged from stochastic numeric label assignments during episodic task sampling. Questioning what defines ``true" meta-learning, we introduce the ``any-way" learning paradigm, an innovative model training approach that liberates model from
fixed cardinality constraints. Surprisingly, this model not only matches but often outperforms traditional fixed-way models in terms of performance, convergence speed, and stability. This disrupts established notions
about domain generalization. Furthermore, we argue that the inherent 
label equivalence naturally lacks semantic information. To bridge this 
semantic information gap arising from label equivalence, we further propose a mechanism for infusing semantic class information into the model. This would enhance the model's comprehension and functionality. Experiments conducted on renowned architectures like MAML and ProtoNet affirm the effectiveness of our method.
\end{abstract}

\section{Introduction}
Meta-learning, often referred to as `learning to learn', is a training strategy designed to enable rapid adaptation to new tasks with only a few examples. Despite the impressive performance of deep learning models on extensive datasets, they still fall short of human abilities when it comes to learning from a small number of instances. To tackle this, various approaches have been proposed \cite{labelequiv:metadefinition}.

\begin{figure}[ht!]
\centering
\includegraphics[width=0.4\textwidth]{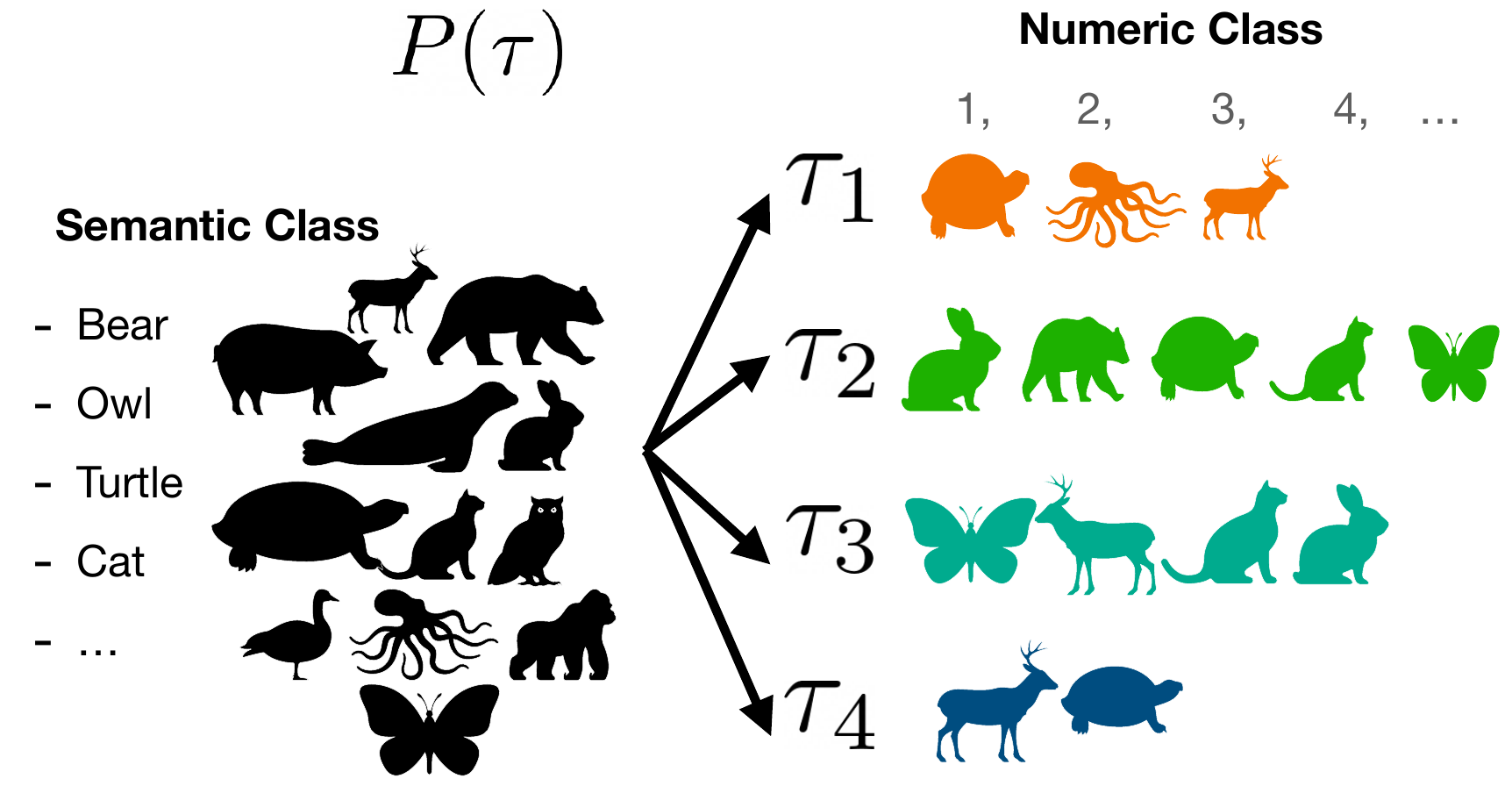}
\caption{Task sampling procedure in our \textit{any-way} meta-learning. Meta-learning employs episodic learning using a mother dataset with numerous classes ($M$) where each label represents each \textit{semantic class} (e.g. `turtle', `octopus', and `deer'). Each episode samples $N < M$ classes from this dataset and assigns a random \textit{numeric class} ($\in [N]$) to each selected class. The flexibility of our \textit{any-way} setting allows varying $N$ across tasks, distinguishing it from the conventional fixed sampling procedure.}
\label{fig:task_sampling}
\end{figure} 

There are two main meta-learning approaches. The first approach focuses on creating an effective feature extractor, like ProtoNet \cite{episodic:MBML:Protonet}, which can accurately cluster unseen instances from the same class to enable the model to generalize well to new data. The second approach aims to find a suitable initialization point for the model, facilitating rapid adaptation to unseen tasks \cite{episodic:GBML:MAML, episodic:GBML:LEO}. 

To imbue the model with meta-learning ability, both methods involve episodes. Each episode consists of multiple tasks and tests the model's meta-learning ability in the regime of few-shot learning
(Fig.\ref{fig:task_sampling}). This continuous exposure to new tasks helps the model improve its learning ability from limited data, simulating the conditions where humans learn.

Nonetheless, in contrast to the human's remarkable flexibility in classifying varying numbers of classes, the current learning models remain rigid, restricted to the fixed `way' they were trained on. As evident in Table~\ref{table:intuition}, when a model trained on 10-way-5-shot tasks from MiniImageNet \cite{dataset:MiniImageNet} is tested on a 3-way-5-shot cars \cite{dataset:cars} setting, its performance plunges to a mere 41$\%$. Considering that random sampling would yield at least 33$\%$, this reveals a glaring deficiency in meta-learning's adaptability across different `way' settings.

The sharp decline in performance is more than just a testament to the model's inflexibility; it also suggests that different `way' episodes might hail from different distributions. This variation between distributions touches upon a deeper, underlying problem: domain generalization. Essentially, to achieve true `any-way' learning, our models must address the broader challenge of performing effectively even in disparate, unseen domains.

The core objective of this paper is to highlight and address this challenge, advocating for an approach where task cardinality agnosticity becomes a foundational element in meta-learning.
To deal with this problem, we propose an \textit{‘any-way’} meta-learning method based on the concept of ‘label equivalence’, which emerges from the task sampling process, an essential element in few-shot learning including meta-learning.
The task sampling process has a unique characteristic: classes are randomly sampled and assigned to (typically numeric) class labels within each episode. These labels are not assigned based on the inherent semantics of the classes; rather, they are determined by the requirements of the specific tasks in each episode. As a result, the same class can be assigned different labels in different tasks and episodes, and the meaning of a label can change from one task to another. As a result, a numeric class label in a given task does not necessarily represent a specific class\footnote{In this paper, we differentiate the `semantic class' from the `numeric class label' as shown in Figure~\ref{fig:task_sampling}.}. This distinction introduces a phenomenon we term `label equivalence', which suggests that all output labels are functionally equivalent to one another. Although there exists a paper \cite{lee2023shot} which argues that each inner loop is equivalent to a whole training trajectory in conventional deep learning, no previous paper has developed this phenomenon in terms of label equivalence.

\begin{table}[t]
\centering
\begin{tabular}{c|cc|cc}
\hline
\# of ways (train)    & \multicolumn{1}{c|}{3} & 10    & \multicolumn{1}{c|}{5} & 10    \\ \hline
\# of ways (test)     & \multicolumn{2}{c|}{3} & \multicolumn{2}{c}{5} \\ \hline \hline
MiniImageNet (5) & 76.27      & 62.45     & 64.92     & 55.2      \\ \hline
Cars (5)        & 59.95      & 41.47     & 47.73     & 32.38    \\ \hline
\end{tabular}
\caption{Performance Comparison in MiniImageNet across different numbers of ways. The numbers in parentheses indicate the number of shots.}
\label{table:intuition}
\end{table}

Examining label equivalence naturally brings up an important question: Given the equivalence of all labels or output nodes,
is it necessary 
to match the number of output nodes to the number of classes in each task? For instance, if a new task consists of three classes, we can select 3 nodes from a pool of, let's say,  
30 output nodes and designate them as numeric labels, optimizing solely for this selected path. This implies that there's no need to rigidly stick to a fixed number of numeric labels, potentially allowing us to further generalize the few-shot learning process. It is from this juncture that our discourse begins.

The primary contribution of this paper is the world-premiere proposition of 
`any-way' few-shot learning, moving beyond the conventional fixed-way few-shot learning. Not only do we extend meta-learning to any-way, but we also provide an efficient, general learning algorithm that achieves performance comparable to or even better 
than fixed-way learning. Moreover, label equivalence further suggests that individual outputs cannot represent the semantic of a class. This arises from the fact that each few-shot task 
contains no semantic information about its class, nor does it recognize or account for the classes it comprises. From this perspective, we can perceive supervised learning as solving a given task within an absolute coordinate system (one axis for one semantic class), while few-shot learning can be viewed as problem-solving within a relative coordinate system, where the uniqueness of each class is the only clue for learning. 

Our second contribution deals with the challenges arising from the lack of semantic class information. In this paper, we introduce an algorithm that compensates for this issue by integrating semantic class information into the learning process. We show that our algorithm not only improves performance in environments where the distribution of the test set differs from that of the training set, but also enables the application of data-augmentation techniques used in supervised learning to few-shot learning tasks. We demonstrate the effectiveness of this approach in MAML and ProtoNet, two of the most representative methods of episode-based meta-learning.

\begin{figure*}[!ht]
\centering
\includegraphics[width=\textwidth]{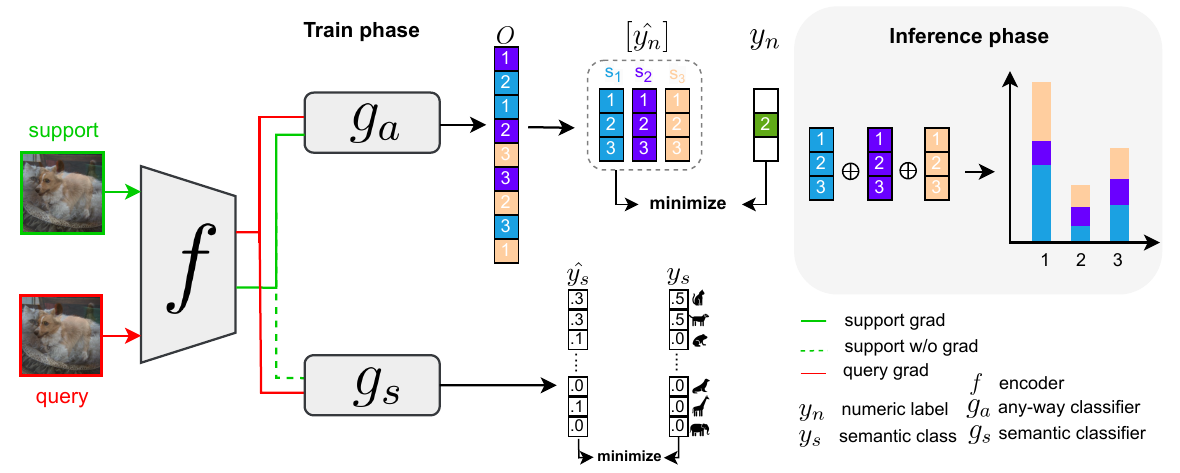}
\caption{Overall structure of our proposed algorithm. The upper section outlines the implementation of the any-way meta-learning, while the lower section highlights the integration of semantic class information. Note that the encoder $f$ is not updated with the semantic classifier $g_s$ during the inner loop (dotted green line). For mixup input, a separate numeric label is assigned in $g_a$, while traditional mixup training is done in $g_s$.}

\label{fig:our_architecture}
\end{figure*}

\section{Related Works}
\paragraph{Meta-Learning,} briefly described, is about the study of algorithms that learn from data. As defined in \cite{labelequiv:metadefinition}, the fundamental concept of meta-learning is the model's capability to discern relationships among a limited number of samples.
Furthermore, mechanisms were proposed in \cite{label_equiv:matchingnetwork} to continuously verify meta-learning abilities while simultaneously enhancing performance.

Inspired by the Matching network \cite{label_equiv:matchingnetwork}, a majority of current meta-learning models have adopted an episode-based training approach. As summarized by \citet{nonepisodic:transferminitocars}, these episode-based models usually consist of two stages: the meta-training stage to acquire meta-learning capabilities, and the meta-test stage to evaluate their performance. There are two lines of research in this area:

\paragraph{Gradient-Based Meta Learning (GBML)} 
 introduced by the Model-Agnostic Meta-learning model (MAML) \cite{episodic:GBML:MAML}, it involves training the support set in the inner loop through gradient descent. The degree of optimization is then evaluated on the query set, followed by the outer loop refining the meta-initialization parameter to enhance learning on the support set. 
There's a growing body of research in GBML
such as \cite{episodic:GBML:ANIL, episodic:GBML:imaml, episodic:GBML:warpgrad, episodic:GBML:Reptile}. Notably, due to the computational demands of caluculating the Hessian in the inner loop, 
methods that utilize latent vectors to reduce this computation 
have been explored \cite{episodic:GBML:LEO}. Other studies have put forward the idea of 
avoiding the use of the Hessian altogether \cite{episodic:GBML:ANIL, episodic:GBML:Reptile, episodic:GBML:MAML}.

\paragraph{Metric-Based Meta Learning (MBML)} is focused on training an effective encoder that can perform clustering at the feature level. This approach, which predates GBML, was established by initiatives such as \cite{label_equiv:matchingnetwork}. 
A foundational architecture for this was proposed by Protonet \cite{episodic:MBML:Protonet}. The method involves extracting features for the inputs in the support set by an encoder, clustering them, and then minimizing the distance between the query set and the corresponding class. Although clustering is conventionally conducted in the Euclidean space, it has been shown to be more effective in other Riemannian spaces like the Hyperbolic space \cite{episodic:MBML:Hyperbolic}.

\paragraph{Domain Generalization} 
This field focuses on designing models that exhibit robustness across diverse environments~\cite{DG:Gunbon0,DG:Gunbon1}. Our approach to ``any-way" meta-learning aligns with this objective. As illustrated in Table~\ref{table:intuition}, episodes of varying cardinality can be perceived as from distinct distributions. In pursuing a model endowed with a genuine meta-learning capability— capable of adapting to any task cardinality— we inevitably confront the challenges posed by domain generalization.

The conventional understanding suggests that while it is possible to enhance the overall performance on a set of tasks, the performance on individual datasets might decline 
when compared to models trained exclusively 
on a specific episode~\cite{DG:wisdom0,DG:wisdom1,DG:wisdom2}. In this paper, we challenge this established notion. 
Through the integration of label equivalence, we posit that this trade-off might not be as significant 
within the meta-learning realm. Our results exhibit not only the seamless adaptability to ``any-way" tasks but also superior performance compared to fixed-way training paradigms.

\section{Method}
\subsection{Conventional (Fixed-Way) Meta-Learning}

To evaluate the few-shot learning ability within a given distribution, we define a process of task sampling. Initially, 
we select a label set from the class pool of a larger mother dataset  $\mathcal{D}$. In conventional meta-learning, the cardinality $N$ of the selected label set remains
fixed during the entire training procedure. We then sample data from this set, each matching the size of the support set (number of shots).
Each task $\tau$ consists of a set $\mathcal{X}$ of tuples $\{(x, y)\}$ sampled from $\mathcal{D}$, where $x$ denotes the input data and $y$ represents the numeric label rather than the semantic class from the dataset (Fig. \ref{fig:task_sampling}).  Moreover, $\mathcal{X}$ is divided into a support set $\mathcal{X}_s$ and a query set $\mathcal{X}_q$ for task learning and performance evaluation, respectively. 
To do so, a model with encoder $f$ and $N$-way classifier $g_{f}$\footnote{$f$ denotes `fixed-way' while $a$ is used for `any-way'. } learns from the support set, then evaluates its performance with the query set.

\begin{algorithm*}
\caption{Any-Way Meta-Learning}\label{alg:AnyWayMetaLearning}
\begin{algorithmic}

\Require Output dimension $O$ 
\Ensure Trained model $\theta$
\State Initialize model parameters $\theta$
\For{episode $= 1$ to $K$}
    \State Sample a random number $N \le O$
    \State Sample a task $\tau$ with $N$ number of classes and set $J = \lfloor O/N \rfloor$.
    \State Generate $S = \{s_j\}, j \in [J]$ where $s_j \in \mathbb{R}^N$ is a random assignment, non-overlapping with $s_i, \ \forall i\neq j$ 
    \State Calculate any-way loss in the inner loop: $L_{a\_in} = \sum_{j=1}^J L(S_j(g_a(f(x_{support})), \theta),y)$ 
    \State update  $\theta$ using $L_{a\_in}$ to $\hat{\theta}$
    \State Calculate any-way loss in the outer loop: $L_{a\_out}$ $=\sum_{j=1}^J L(S_j(g_a(f(x_{query})), \hat{\theta}),y)$
    \State Update meta-parameters $\theta$ using $L_{a\_out}$.
\EndFor
\State \textbf{return} trained model with parameters $\theta$
\Ensure Test a model with trained model $\theta$
\State Sample a task $\tau$ from test pool and set $J = \lfloor O/N \rfloor$ and Generate $S$ in same manner. Then , update $\theta$ to $\hat{\theta}$ with $L_{a\_in}$
\State Calculate ensembled logit: $logit$ = $\sum_{j=1}^J S_j(g_a(f(x_{query})))$ to achieve test accuracy
\end{algorithmic}
\end{algorithm*}

\subsection{Any-Way Meta-Learning}\label{sec:anywaymetalearning}
As mentioned earlier, there exists an equivalence among numeric labels. 
We can leverage
these characteristics to enhance the meta-ability of the model.


Firstly, our model's output dimension, $O$, is set to be greater than the maximum cardinality of the tasks' label sets, implying that each time we sample a task $\tau_i$ with $N_i$ classes, we must assign $N_i$ labels among $O$ output nodes. 

With our proposed configuration, we maintain output-to-class assignments during each episode. This not only determines a unique and unchanging pathway for selecting numeric labels but also restricts gradient updates to occur exclusively along this fixed way, thereby accommodating the execution of standard meta-learning tasks. Unlike traditional methods that fix the cardinality of the task when conducting task sampling, our approach also involves sampling the cardinality. 
This enables our model to handle tasks with any cardinality, hence fostering a cardinality-independent learning environment, \ie any-way meta-learning. 

However, a potential problem can arise when the output dimensionality $O$ significantly exceeds the expected task cardinality. In this case, the chance of each classifier node being selected 
in each episode approaches zero. Consequently, classifier nodes might undergo less training than the encoder throughout the training procedure, resulting in an underfit.
To remedy this, we assign a set of output nodes to a numeric label. We assign a numeric label to (almost) all nodes for classification, rather than assigning a numeric label to just one output node and ignoring the unassigned nodes.

Let $f(x)$ be a feature map for the input $x$ and $g_a(\cdot)$ be an any-way classifier consisting of $O$ output nodes. For a task with $N$ classes, we can constitute \nj{a set of $J = \lfloor O/N \rfloor$ assignments $S = \{s_j \in \mathbb{R}^N$, $j \in [J]\}$, in which the $i$-th element of $s_j$} corresponds to the output node for the $i$-th numeric class label\footnote{For example, consider a scenario with eight classifier output nodes and three classes in a task, thus, three numeric labels. In this case, it can be $s_1 = [3,5,2]^T$, $s_2 = [7,4,8]^T$ and nodes 1 and 6 are unassigned. }.  Here, $\lfloor \cdot \rfloor$ is the floor operation, and $[n]$ denotes a set of natural numbers from $1$ to $n$. We then aggregate the loss for each assignment to constitute our any-way loss $L_a$ as
\begin{equation}
L_{a} = \sum_{j=1}^J L(S_j(g_a(f(x))),y).
\label{eq:any-way-loss}
\end{equation}
Here, $S_j: \mathbb{R}^O \rightarrow \mathbb{R}^N$ uses the vector $s_j$ as an index function to extract $N$ elements, \eg $S_1(v)$ with $s_1 = [3,5,2]^T$ outputs $[v_3, v_5, v_2]^T$.    As different assignments do not overlap, \ie $\text{set}(s_i) \ \cap \ \text{set}(s_j) = \phi$ where $\text{set}(v)$ is the set consisting of elements of the  vector $v$, this additional assignment procedure 
requires no additional computation. \ie no additional backpropagation nor inference.

\subsection{Injecting Semantic Class Information To Model}
While it's possible to implement any-way meta-learning through label equivalence, \jh{the fact that} relying solely on numeric labels \jh{still} leaves out important semantic information \jh{as previously mentioned}, is a significant limitation. 
\jh{In this section, we aim to remedy this problem}. To address this issue, we incorporate an auxiliary classifier subsequent to the encoder, effectively bridging the gap.
This auxiliary classifier becomes semantic classifier as it classifies semantic labels.
The overall structure is illustrated at the bottom of Fig.~\ref{fig:our_architecture} ($g_s$).


To ensure consistency and simplicity in our approach, the encoder is kept unchanged throughout the process. This allows for a streamlined, single-pass operation through the network. Specifically, the semantic classifier receives features from the encoder, which operates with meta-initialized parameters. Importantly, this means that the features extracted by the encoder remain constant during the inner loop, ensuring stability and reliability in the feature extraction process. 
For instance, consider training on the TieredImageNet dataset. This benchmark includes a total of 608 classes in its training dataset. 
Accordingly, the semantic classifier should have 608 outputs to match these classes.

\begin{table*}[]
\centering
\begin{tabular}{l|ccc|ccc|ccc|ccc}
\toprule
Dataset & \multicolumn{3}{c|}{MiniImageNet} & \multicolumn{3}{c|}{TieredImageNet} & \multicolumn{3}{c|}{Cars} & \multicolumn{3}{c}{CUB} \\ 
\midrule
Num-Way & 3 & 5 & 10 & 3 & 5 & 10 & 3 & 5 & 10 & 3 & 5 & 10 \\ 
\midrule
f-MAML (1) acc& 59.56 & 46.86 & 31.91 & 57.33 & 46.76 & 33.88 & 64.02 & 49.19 & 32.84 & 70.58 & 57.35 & 41.32 \\ 
\hfill std  & 1.48 & 0.95 & 0.50 & 0.82 & 0.58 & 0.43 & 0.95 & 0.56 & 1.91 & 0.85 & 0.58 & 0.98 \\ 
\midrule
a-MAML (1) acc& 63.15 & 48.33 & 31.34 & 61.92 & 47.43 & 31.62 & 64.72 & 50.24 & 33.79 & 72.21 & 58.85 & 41.66 \\ 
\hfill std  & 0.52 & 0.26 & 0.28 & 0.26 & 0.28 & 0.18 & 0.24 & 0.23 & 0.26 & 0.19 & 0.33 & 0.42 \\
\midrule
f-MAML (5) acc & 76.64 & 65.41 & 51.18 & 73.16 & 67.69 & 47.96 & 79.54 & 66.59 & 50.05 & 83.73 & 74.71 & 60.45 \\ 
\hfill std  & 0.41 & 1.04 & 0.27 & 0.66 & 0.23 & 0.37 & 0.85 & 1.22 & 1.38 & 1.16 & 0.40 & 1.02 \\ 
\midrule
a-MAML (5) acc& 79.06 & 66.73 & 50.28 & 79.30 & 68.10 & 52.19 & 81.55 & 70.42 & 52.67 & 84.69 & 75.20 & 59.81 \\ 
\hfill std  & 0.38 & 0.42 & 0.49 & 0.47 & 0.63 & 0.78 & 0.49 & 0.88 & 2.23 & 0.10 & 0.44 & 0.86 \\
\bottomrule
\end{tabular}
\caption{Few-shot classification across various task cardinalities. The training and testing were conducted within the same benchmark with 4-conv. While implementing 10-way-1-shot f-MAML, we encountered many cases of corruption, \ie we failed to train the model. We discarded all these failed experiments and sampled three `successful' models with different seeds for the performance report. For our a-MAML with $O = 30$, we did not encounter any corruption.}
\label{table:GBML_MAIN}
\end{table*}

In our implementation, the semantic loss, $L_{semantic}$, which is a typical cross-entropy loss with $C (\gg N)$ classes, is added to the original loss (either a fixed-way loss or an any-way loss in (\ref{eq:any-way-loss})) using a balancing weight $\lambda$, \ie $L_{total} = L_{original} + L_{semantic}$. It's important to note that the semantic labels encountered in the test phase are completely unseen during the training phase. This suggests that any improvement in performance is not merely a result of simple overfitting.

The inclusion of the semantic classifier provides a gateway to information that was 
inaccessible 
in conventional episode-based meta-learning. This unique structure, where supervised learning is inherent in the semantic classifier in every episode, paves the way for incorporating conventional supervised learning techniques into our approach.
As an exemplar, we employ the Mixup technique, one of the most commonly used data-augmentation techniques in supervised learning. Using the known semantic class\nj{es}, we apply this technique as it is used in supervised learning. 
While the numeric labels stay unchanged, we blend the semantic class across tasks, generating new input-output pairings.
This practical implementation of the Mixup technique in our methodology emphasizes our central assertion: conventional supervised learning strategies can be seamlessly and effectively integrated within the meta-learning landscape.
\begin{table*}[]
\centering
\begin{tabular}{l|ccc|ccc|ccc|ccc}
\toprule
Scenario & \multicolumn{3}{c|}{G$\rightarrow{}$G} & \multicolumn{3}{c|}{S$\rightarrow{}$S} & \multicolumn{3}{c|}{G$\rightarrow{}$S} & \multicolumn{3}{c}{S$\rightarrow{}$G} \\ 
\midrule
Train $\rightarrow{}$ Test & \multicolumn{3}{c|}{Mini $\rightarrow{}$Tiered} & \multicolumn{3}{c|}{Cars$\rightarrow{}$CUB} & \multicolumn{3}{c|}{Mini$\rightarrow{}$Cars} & \multicolumn{3}{c}{Cars$\rightarrow{}$Mini} \\ 
\midrule
Num-Way                   & 3 & 5 & 10 & 3 & 5 & 10 & 3 & 5 & 10 & 3 & 5 & 10 \\ 
\midrule
f-MAML (1) acc  & 61.86 & 50.78 & 36.62 & 45.24 & 31.38 & 18.13 & 49.29 & 35.09 & 22.06 & 43.42 & 28.76 & 15.95 \\
\hfill std  & 1.29 & 0.77 & 0.12 & 1.18 & 0.28 & 1.26 & 0.73 & 0.66 & 0.11 & 0.91 & 0.91 & 1.04 \\
\midrule
a-MAML (1) acc  & 66.27 & 52.27 & 35.62 & 46.92 & 32.01 & 17.72 & 49.51 & 35.22 & 21.36 & 43.66 & 28.64 & 15.58 \\
\hfill std  & 0.36 & 0.25 & 0.20 & 0.17 & 0.14 & 0.62 & 0.04 & 0.28 & 0.07 & 0.20 & 0.08 & 0.30 \\
\midrule
f-MAML (5) acc  & 78.70 & 68.54 & 54.64 & 56.57 & 43.60 & 27.35 & 61.58 & 48.07 & 34.67 & 51.91 & 39.41 & 24.76 \\
\hfill std  & 0.37 & 1.17 & 0.34 & 0.30 & 1.40 & 0.59 & 1.53 & 1.83 & 0.67 & 0.86 & 0.62 & 0.81 \\
\midrule
a-MAML (5) acc  & 80.46 & 69.73 & 53.68 & 62.04 & 45.63 & 27.02 & 63.58 & 49.80 & 34.82 & 57.29 & 41.13 & 24.15 \\
\hfill std  & 0.17 & 0.17 & 0.19 & 0.97 & 1.58 & 2.47 & 0.72 & 0.83 & 0.96 & 0.55 & 0.79 & 1.13 \\
\bottomrule
\end{tabular}
\caption{Few-shot classification across various task cardinalities using 4-conv. We tested the generalizability of our model across various scenarios. Here, G: General, S: Specific, Mini: MiniImageNet, Tiered: TieredImageNet}
\label{table:GBML_CROSS}
\end{table*}

\section{Experiment}
\subsection{Implementation Details}
\paragraph{Datasets} We evaluated our methodology on a diverse range of datasets. The general datasets, denoted as `G', such as MiniImageNet \cite{dataset:MiniImageNet} and TieredImageNet \cite{dataset:Tieredimagenet} (subsets of the more extensive ImageNet \cite{dataset:imagenet} with 100 and 400 classes respectively) serve as versatile bases for broader tasks. In contrast, the Cars \cite{dataset:cars} and CUB \cite{dataset:cub} datasets, representing specific datasets or `S', are widely used for fine-grained image classification evaluations due to their focus on closely related objects with subtle variations. 
By utilizing these datasets, we are able to comprehensively evaluate the performance of our methodology across different task spectrums. 

\paragraph{Task Sampling} For each episode's initiation, the task cardinality was randomly sampled from \{3, 5, 7, 9\}. While tests were 
conducted on 3-way, 5-way, and 10-way, the 10-way cardinality was excluded from our sampling pool.

\paragraph{Environments} We implemented MAML and ProtoNet using torchmeta library \cite{coding:torchmeta}, using singe A100 GPU.

\paragraph{Hyperparameters} In line with \cite{episodic:GBML:BOIL}, our experiments involved sampling 60,000 episodes. We adopted the 4-conv architecture as detailed in \cite{dataset:MiniImageNet}. The learning rates were set at 0.5 for the inner loop and 0.001 for the outer loop. Depending on the shot type, the $\lambda$ values were adjusted: 0.1 for 1-shot and 0.5 for 5-shot for MAML. And 0.1 for 5-shot and 0.01 for 1-shot in ProtoNet. For the mixup showcase, we followed the convention of sampling the mixup rate from a beta distribution with $\alpha = \beta = 0.5$.  
When adopting mixup, we assign a separate numeric label to the mixed inputs.
 Also, when constructing prototye vectors, the EMA (exponential moving average) rate was set to 0.05 for 5-shot and 0.01 for 1-shot.

\paragraph{Algorithms} Our primary experiment was grounded in MAML due to its inherent generalized structure, enabling seamless adaptability and 
broad generalizability. We also conducted experiments using Protonet \cite{episodic:MBML:Protonet}, a well-acknowledged method within MBML. Demonstrating efficacy in both GBML and MAML reinforced our propositions.
We evaluated our model which scored the best validation accuracy. Given that we sample the task cardinality, we simply calculated the validation accuracy by summing the validation accuracy across all task cardinalities. 

\paragraph{Scenarios} As articulated in \cite{nonepisodic:transferminitocars}, the merit of meta-learning extends beyond achieving high performance  
within the trained distribution. It's equally pivotal to maintain competent performance when the test distribution deviates from the training paradigm. Therefore, our experimental designs spanned four distinct cross-adaptation scenarios: transfer within general datasets (G$\rightarrow$G), transfer from general to specific datasets (G$\rightarrow$S), transfer from specific to general datasets (S$\rightarrow$G), and transfer within specific datasets (S$\rightarrow$S).

\paragraph{Expansion to Metric-Based Learning} 
Initially, our study centered around models
equipped with classifiers capable of 
discerning semantic classes. However, since metric-based models lack such classifiers, we recognized the need for refining our algorithm. 
This challenge steered us towards seminal research in the realm of continual learning. Noteworthy contributions, as highlighted by \cite{continual:catastrophic,catastrophygunbon,continual:memoryrehearsal}, elucidate key techniques to mitigate the `catastrophic forgetting' dilemma — a phenomenon where models unintentionally discard previously assimilated knowledge during subsequent learning.
Guided by these insights, we constructed a scheme centered on crafting a list of prototype vectors, each representing 
a distinct semantic class. Throughout the lifecycle of every episode, we optimize the model and reduce the disparity between the newly minted prototype vector and its semantically aligned counterpart. Also, we update semantic prototypes via EMA to reflect the model's changing over time. This cyclical mechanism strengthens our model's capability to seamlessly preserve and leverage information of each semantic class. The metric elineating the discrepancy between the semantic and prototype vectors within each episode has been incorporated as a crucial balancing factor, mirroring its role in $L_\text{semantic}$.

\subsection{Any-Way Meta Learning vs. Fixed-Way Methods}

As shown in Tables.~\ref{table:GBML_MAIN} and \ref{table:GBML_CROSS} our any-way method outperforms conventional fixed-way method in most benchmarks. This is surprising because one might conventionally assume that fixed-way methods would be specialized for a specific episode distribution. However, the superiority of our approach, which remains agnostic to task-cardinality, challenges this conventional assumption.

One might argue that the improved performance is due to the exposure to a larger support set, as seen in 9-way episodes, even though the number of episodes remains constant. However, this perspective is refuted by the elevated performance observed in 10-way tasks, which was not part of our training scenarios (\ie unseen distribution; \nj{the training was done with $\{3,5,7,9\}$ classes}).
Considering that the average task cardinality is 6, one would expect superior performance when trained exclusively on 10-way episodes. Nevertheless, a-MAML remains comparable to the baseline. 

This result showcases the strength of our model. Even when trained on a mixed distribution (\ie multiple task cardinalities), it outperforms fixed-way models tailored to \nj{a} specific domain. 
Moreover, it demonstrates comparable or even superior performance compared to fixed-way MAML models trained on unseen task cardinalities. Typically, models are vulnerable to biases from their specific training datasets, which can undermine their performance on new datasets. However, our findings challenge this conventional understanding \cite{theory:specificdataset}.
Thus, our results not only emphasize way-agnositc meta-learning but also challenge this accepted belief in Domain Generalization, \nj{which can be an interesting topic of further research}.
\begin{table*}[]
\resizebox{\textwidth}{!}{%
\begin{tabular}{l|cc|cc|cc|cc}
\multicolumn{1}{c|}{Scenario} & 
  \multicolumn{2}{c|}{Same} &
  \multicolumn{2}{c|}{Same} &
  \multicolumn{2}{c|}{G $\rightarrow$ S} &
  \multicolumn{2}{c}{S $\rightarrow$ G} \\ \hline
\multicolumn{1}{c|}{Train $\rightarrow$ Test} & 
  \multicolumn{2}{c|}{mini $\rightarrow$ mini} &
  \multicolumn{2}{c|}{cars $\rightarrow$ cars} &
  \multicolumn{2}{c|}{mini $\rightarrow$ cars} &
  \multicolumn{2}{c}{cars $\rightarrow$ mini} \\ \hline
\multicolumn{1}{c|}{Num-Way}      & 5 & 10 & 5 & 10 & 5 & 10 & 5 & 10 \\ \hline
ProtoNet (1) &44.38 (1.59)  &28.85 (1.22) & 37.72 (0.62)  & 23.85 (0.33)  &   31.93 (0.73) &  19.72 (0.78)  &27.47 (0.71)   & 15.76 (0.32)   \\
\multicolumn{1}{r|}{+ semantic}    &  44.61 (0.15) & 29.03 (0.02)   &  40.60 (0.09) & 26.07 (0.15)   & 31.26 (0.12)  &  19.20 (0.01)  & 27.93 (1.15)  &  15.91 (0.85)  \\
\multicolumn{1}{r|}{+ mixup}      &  45.43 (0.60) &  29.66 (0.54)  & 38.78 (1.55) & 24.86 (1.35)   & 32.03 (0.39)  & 19.90 (0.38)   & 27.29 (0.66)  & 15.56 (0.48)  \\
a-MAML (1)   & 46.86 (0.95) & 31.91 (0.50)   & 50.24 (0.23)  &  33.79 (0.26)  & 35.22 (0.28)  & 21.36 (0.07) &  28.64 (0.08) & 15.58 (0.30)   \\
\multicolumn{1}{r|}{+ semantic}    & 49.16 (0.95) & 32.28 (0.87)   & 50.93 (0.89)  & 34.04 (0.33) & 34.80 (0.32)  & 22.01 (0.18)   & 29.60 (0.81)  & 15.96 (0.41)   \\
\multicolumn{1}{r|}{+ mixup}     &48.45 (0.35)& 31.62 (0.44)& 53.13 (0.82)  & 36.44 (0.64)          & 35.00 (0.23)  & 22.02 (0.35)   &  29.30 (0.67) & 14.70 (1.17)   \\ \hline
ProtoNet (5) & 59.19 (0.39)  &  43.63 (0.31)  & 53.10 (1.30)  & 38.13 (1.15)   & 42.80 (0.80)  & 29.02 (0.65)   & 36.15 (1.24)  &  22.56 (1.02)  \\
\multicolumn{1}{r|}{+ semantic} & 60.75 (1.38)  &  44.38 (1.61)  &  55.66 (0.32) & 40.66 (0.26)   & 43.93 (0.10)  & 29.86 (0.10)   & 39.41 (1.88)  & 25.09 (1.53)   \\
\multicolumn{1}{r|}{+ mixup}    &  60.76 (1.18) & 44.31 (1.51)   & 58.28 (0.67)  & 42.83 (0.70)   & 42.41 (0.88)  & 28.56 (0.94)   & 38.54 (0.64)  & 24.52 (0.55)   \\
a-MAML (5)   & 66.73 (0.42)  &  50.28 (0.49)  & 70.42 (0.88)  & 52.67 (2.23)   & 49.80 (0.83)  &  34.82 (0.96)  & 41.13 (0.79)  & 24.15 (1.13)   \\
\multicolumn{1}{r|}{+ semantic}    & 66.91 (0.10)  & 50.16 (0.06)   & 70.56 (0.12)  & 52.12 (0.78)   & 49.06 (0.71)   &  34.07 (0.78)  & 40.88 (1.44)  & 23.45 (1.47)   \\
\multicolumn{1}{r|}{+ mixup} &   66.79 (0.07) &  50.05 (0.03) & 70.99 (0.27)   & 53.55 (0.63)  & 50.04 (0.15)   & 34.90 (0.08)  &  43.00 (0.14)  &   25.33 (0.26)   \\ 
\end{tabular}
}
\caption{Evaluating the impact of injecting semantic information and mixup technique in various scenarios on any-way-meta learning and MBML (Metric-Based-Meta-Learning) \cite{episodic:MBML:Protonet}. Numbers in parentheses indicate the shot count, and mini refers to MiniImageNet. Note that the class in the test phase is totally unseen during training.}
\label{table:anysemantic}
\end{table*}


\begin{figure}[t!]
\centering
\includegraphics[width=0.45\textwidth]{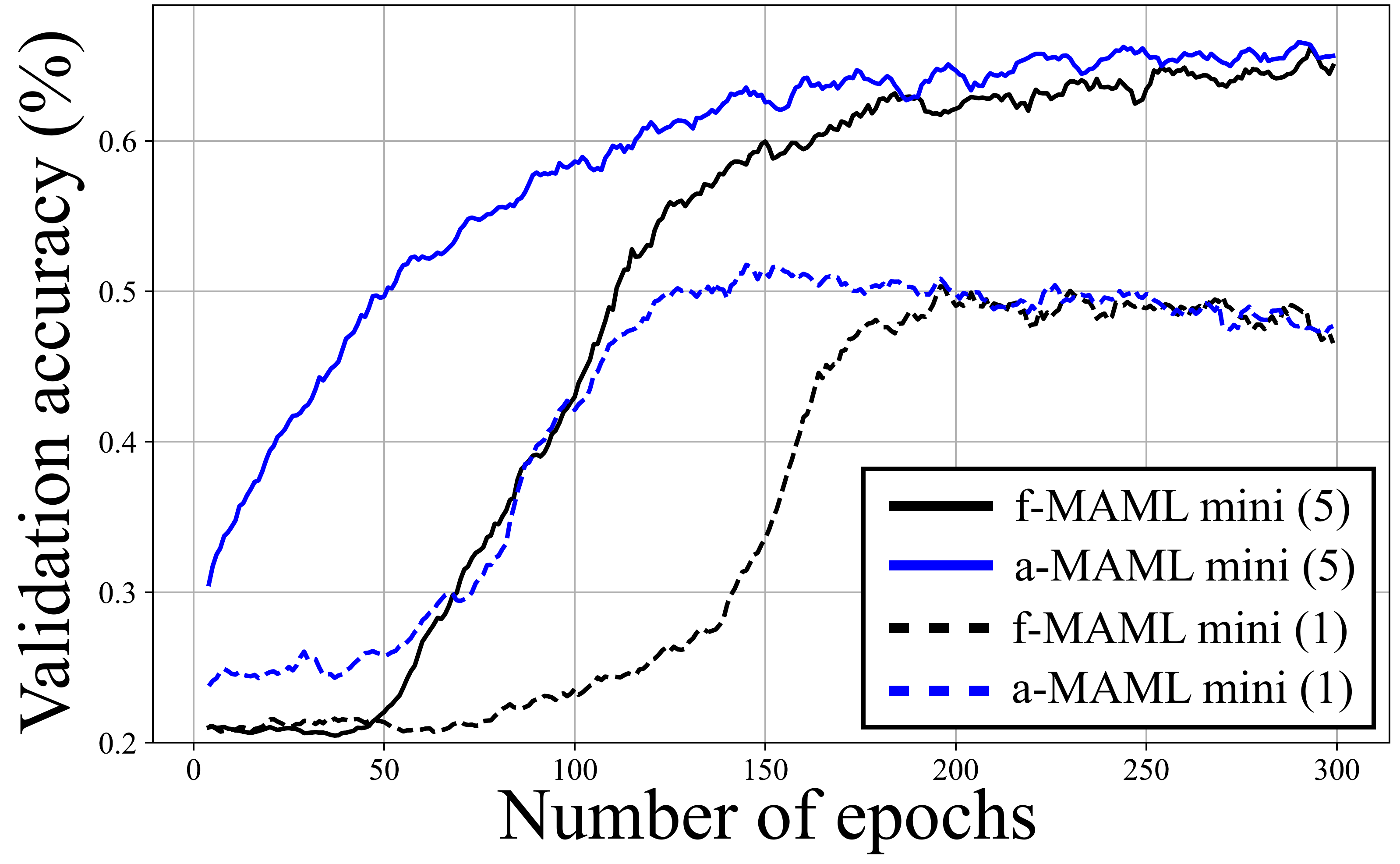}
\caption{A graph comparing validation accuracy upon training epochs. The numbers in parentheses indicate the number of shots. The validation accuracy is measured in the 5-way case. mini refers MiniImageNet.}
\label{fig:convergence_speed}
\end{figure}

\subsection{Any-Way, Through the Lens of Ensemble}
\paragraph{Ensemble on Assigned sets.} Our ensemble-based perspective provides insights into the improved performance of our algorithm. As highlighted in Sec.~III.2, 
the flexibility in assigning numeric labels offers a way to generate diverse models without any additional costs.
By harnessing the multiple logits derived from the same task set, we can utilize 
these `independent' pieces of information.

Table~\ref{table:ensemble1} corroborates this idea. As the number of assignment sets, \nj{$J$,} increases, 
the performance improves due to the model ensembling with logits generated by each distinct set. With our `any-way' approach, it is possible to generate an `almost' infinite number of such assignment sets, each acting as a representation of an independent model. We conducted various basic ensemble methods and all of those methods have shown promising performance improvements as the number of models increases. Nonetheless, there exists a finite limit due to the inherent \nj{upper bound of ensembling}.
In essence, our methodology harnesses the power of ensemble by utilizing varying numeric label assignments. This diversification, when combined, forms a robust model that is more adaptable and delivers superior performance. Additionally, Table~\ref{table:ablation} supports our analysis, showing a positive correlation between the number of output nodes and performance. This is because an increase in the number of outer nodes leads to more ensembling.

\begin{table}[]
\centering
\resizebox{\columnwidth}{!}{%
\begin{tabular}{c|cccccc}

train  & \multicolumn{6}{c}{MiniImageNet}                                                                            \\ \hline
test & \multicolumn{3}{c|}{MiniImageNet}                               & \multicolumn{3}{c}{cars}                   \\ \hline
method        & original     & softmax      & \multicolumn{1}{c|}{max}    & original     & softmax      & max    \\ \hline
1             & 65.00    & 65.00     & \multicolumn{1}{c|}{65.00 }    & 47.66  & 47.66  & 47.66  \\
2             & 66.20   & 66.18  & \multicolumn{1}{c|}{65.11 }  & 49.02  & 49.01  & 47.81  \\
3             & 66.58   & 66.57  & \multicolumn{1}{c|}{66.22 } & 49.31  & 49.3   & 48.86  \\
6             & 66.82 & 66.81  & \multicolumn{1}{c|}{66.54 } & 49.83  & 49.81  & 49.48   \\
12 $\star$           & 66.99  & 66.97  & \multicolumn{1}{c|}{66.69 } & 49.81  & 49.82  & 49.48  \\
18 $\star$           & 66.82  & 66.81  & \multicolumn{1}{c|}{66.59 } & 49.98  & 49.95  & 49.59 
\end{tabular}
}
\caption{Accuracy of a-MAML across various datasets, shots, and ensembling techniques on the 5-way setting. Ensembling methods include: (i) \textit{original}: direct summation of output logits, (ii) \textit{softmax}: \nj{summation of} softmax \nj{outputs} on logits, and (iii) \textit{max}: selection of the maximum output logit per assignment set. The $\star$ refers \nj{to} multiple inner loop\nj{s} for \nj{a} single task for more \nj{ensembling}. The model was trained with 5 shots.}
\label{table:ensemble1}
\end{table}

\paragraph{Ensemble on Different Cardinality Episodes.}
In Fig.~\ref{fig:convergence_speed}, it is evident that our algorithm, a-MAML, converges more rapidly than f-MAML. Beyond its faster convergence, a-MAML also demonstrates superior performance in comparison to f-MAML. This distinction provides insights into the challenges faced during the  training of fixed 10-way 1-shot TieredImageNet \nj{(See the caption of Table \ref{table:GBML_MAIN})}. Specifically, as the complexity of the task increases, the duration of the ``Start-up Stall" phase — where performance remains stagnant after initialization — becomes more pronounced. Given that TieredimageNet is one of the most challenging datasets, primarily due to its expansive semantic class pool (361) and the intricate nature of the task (classifying unseen numeric labels within a single shot), the advantages of a-MAML become even more apparent. By ensembling across different cardinalities, a-MAML swiftly moves past the Start-up Stall phase during simpler tasks like 3-way episodes. This proficiency with easier tasks subsequently empowers the algorithm to tackle harder tasks, such as 9-way episodes, more effectively.

\subsection{Injecting  Semantic information into model}
In our experiments, as detailed in Tables \ref{table:anysemantic} and \ref{table:fixedsemantic}, we found that the integration of semantic class information significantly enhanced the performance of models, especially when tailored for fine-grained datasets. However, the effects of this integration varied depending on the characteristics of the datasets \jh{in any-way case}. 
For models trained on coarse-grained datasets \nj{(G$\rightarrow$S)}, there was a potential decrease in generalizability, as evidenced by diminished performance on intricate datasets like ``MiniImageNet". In contrast, models trained on detailed datasets, such as ``Cars" or ``CUB", not only maintained but even demonstrated enhanced versatility across varied test scenarios. This suggests that the subtle intricacies 
of fine-grained datasets are further emphasized 
with semantic nuances, enabling models to discern delicate 
class differences with greater precision.

\paragraph{Mixup: Crafting Geometry into feature embeddings.} 
\yr{While it is generally beneficial} to inject semantic class {information}, there are \yr{some instances where performance degradation is observed.} 
\yr{This occurs due to the inherent nature of features, which do not readily adapt to geometry. Instead, they lie in a smaller manifold, resulting in intricate challenges when capturing their metrics. To mitigate these cases, we could implement techniques commonly used in supervised learning, specifically Mixup, as shown in Table~\ref{table:anysemantic}.}

\yr{Previous} research~\cite{scarcitylabel} has already \yr{employed Mixup} for meta-learning \yr{objectives}. However, \yr{such implementations primarily address data scarcity by increasing the number of shots within a single episode through Mixup.}
In contrast, our approach \yr{focuses on the} 
the geometry of the feature space without leveraging \yr{additional} 
shots in individual tasks. As we did not leverage the benefits of additional shots by Mixup, we can apply other mixup methods designed to address dataset scarcity simultaneously; that is, we can anticipate further performance enhancement. 
\begin{table}[t]
\centering
\adjustbox{width=0.40\textwidth}{
\begin{tabular}{c|cc|cc|cc} 
{Way} & \multicolumn{2}{c|}{3} & \multicolumn{2}{c|}{5} & \multicolumn{2}{c}{10} \\
\hline

{O \textbackslash  Data} & {mini}  & {cars}  & {mini}  & {cars}  & {mini}  & {cars}  \\ \hline
10       & 76.79         & 60.53 & 63.91         & 46.04 & 46.34         & 30.62  \\
20       & 78.83         & 62.01 & 66.71         & 47.92 & 49.96         & 32.71  \\
30       & 79.06         & 63.58 & 66.73         & 49.80  & 50.28         & 34.82  \\
\end{tabular}
}
\caption{Experiments on the number of output nodes $O$. A model was trained on the Mini-ImageNet dataset using an any-way training method, excluding the semantic classifier.}
\label{table:ablation}
\end{table}

Our auxiliary classifier $g_s$ classifies mix-up labels at the same ratio\yr{, thereby reflecting the model's} disentangled feature space \yr{corresponding to} mix ratios in the input spaces. \yr{This enables} the model to adopt a geometric invariance. By increasing such invariance, the model enhances its \nj{generalizability} \yr{while} maintaining its structure.

Table~\ref{table:anysemantic} demonstrates the overall performance improvement achieved with Mixup. However, in some cases \nj{especially} in `Same' scenarios, performance degradation occurs due to the 
trade-off between generality and specificity \cite{theory:specificdataset}. This arises from the fact that, while Mixup enables the adoption of general features, it also leads to the loss of specific features that could be advantageous when applied within the same domain. 

\section{Conclusion and Future Work}
In this work, we primarily proposed a generalization problem in meta-learning across various task cardinalities. To remedy this issue, we  bring the innate characteristic of episode-based learning, label equivalence. By harnessing label equivalence, we could implement a model capable of dealing with any task cardinality. Our any-way meta-learning surpasses fixed-meta learning, which goes against conventional wisdom. Additionally, we claimed that this equivalence naturally lacks semantic information to classes, thereby we devised a method to inject semantic information into the model. Moreover, our proposed model exhibits some characteristics of supervised learning, prompting us to explore the potential of integrating supervised learning techniques. 
We envision our paper as a catalyst for raising essential inquiries: `What truly defines meta-learning ability?' and `What unfolds when models are trained with episode-sampled data?'

Our algorithm is intuitive and straightforward, utilizing a fundamental and general architecture for broad applicability. 
Additionally, we employed the simplest 
ensemble technique, as seen in Table~\ref{table:ensemble1}. As part of our ongoing efforts, we are actively developing more advanced algorithms to harness the potential of this label equivalence and to further enhance the ensemble technique enabled by this equivalence.
\begin{table}[t]
\centering
\resizebox{0.95\columnwidth}{!}{%
\begin{tabular}{l|l|l|l|l}
\multicolumn{1}{c|}{Scenario}   & \multicolumn{1}{c|}{Same}              & \multicolumn{1}{c|}{Same}              & \multicolumn{1}{c|}{G $\rightarrow$ S} & \multicolumn{1}{c}{S $\rightarrow$ G}   \\ \hline
\multicolumn{1}{c|}{Train$\rightarrow$Test} & \multicolumn{1}{c|}{M $\rightarrow$ M} & \multicolumn{1}{c|}{C $\rightarrow$ C} & \multicolumn{1}{c|}{M $\rightarrow$ C} & \multicolumn{1}{c}{C $\rightarrow$ M} \\ \hline
\multicolumn{1}{c|}{f-MAML (1)} & 46.86 (0.95)      & 49.19 (0.56)      & 35.09 (0.66)      & 28.76 (0.91)      \\
\multicolumn{1}{l|}{+semantic}  & 48.32 (0.89)      & 51.21 (0.36)       & 34.65 (0.20)      & 29.48 (0.36)      \\ \hline
\multicolumn{1}{c|}{f-MAML (5)} & 65.41 (1.04)      & 70.42 (0.88)      & 48.07 (1.83)      & 39.41 (0.62)      \\
\multicolumn{1}{l|}{+semantic}  & 66.89 (0.35)      & 71.45 (1.10)      & 46.19 (2.76)      & 40.46 (0.61)     
\end{tabular}
}
\caption{Testing the effectiveness of injecting semantic information to fixed-way meta-learning. The \nj{numbers in parentheses are the number of shots}. M and C refer \nj{to} MiniImageNet and Cars, \nj{respectively. `+semantic' indicates a model trained with an additional semantic classifier with the $L_{semantic}$ loss. } Note that the class in the test phase is totally unseen during training.} 
\label{table:fixedsemantic}
\end{table}

\section{Acknowledgements}
This work was supported by IITP grants (2022-0-00953, 2021-0-01343) and NRF grant (2021R1A2C3006659), all funded by MSIT of the Korean Government.

\clearpage

\bibliography{reference/aaai24}

\begin{thebibliography}{29}
\providecommand{\natexlab}[1]{#1}

\bibitem[{Baxter(2011)}]{DG:wisdom0}
Baxter, J. 2011.
\newblock A Model of Inductive Bias Learning.
\newblock \emph{CoRR}, abs/1106.0245.

\bibitem[{Caruana(1997)}]{DG:wisdom1}
Caruana, R. 1997.
\newblock Multitask Learning.
\newblock \emph{Machine Learning}, 28(1): 41--75.

\bibitem[{Chen et~al.(2019)Chen, Liu, Kira, Wang, and Huang}]{nonepisodic:transferminitocars}
Chen, W.; Liu, Y.; Kira, Z.; Wang, Y.~F.; and Huang, J. 2019.
\newblock A Closer Look at Few-shot Classification.
\newblock \emph{CoRR}, abs/1904.04232.

\bibitem[{Deleu et~al.(2019)Deleu, W\"urfl, Samiei, Cohen, and Bengio}]{coding:torchmeta}
Deleu, T.; W\"urfl, T.; Samiei, M.; Cohen, J.~P.; and Bengio, Y. 2019.
\newblock {Torchmeta: A Meta-Learning library for PyTorch}.
\newblock Available at: https://github.com/tristandeleu/pytorch-meta.

\bibitem[{Doersch, Gupta, and Efros(2015)}]{label_equiv:matchingnetwork}
Doersch, C.; Gupta, A.; and Efros, A.~A. 2015.
\newblock Unsupervised Visual Representation Learning by Context Prediction.
\newblock \emph{CoRR}, abs/1505.05192.

\bibitem[{Finn, Abbeel, and Levine(2017)}]{episodic:GBML:MAML}
Finn, C.; Abbeel, P.; and Levine, S. 2017.
\newblock Model-Agnostic Meta-Learning for Fast Adaptation of Deep Networks.
\newblock \emph{CoRR}, abs/1703.03400.

\bibitem[{Flennerhag et~al.(2019)Flennerhag, Rusu, Pascanu, Visin, Yin, and Hadsell}]{episodic:GBML:warpgrad}
Flennerhag, S.; Rusu, A.~A.; Pascanu, R.; Visin, F.; Yin, H.; and Hadsell, R. 2019.
\newblock Meta-learning with warped gradient descent.
\newblock \emph{arXiv preprint arXiv:1909.00025}.

\bibitem[{Ghifary et~al.(2015)Ghifary, Balduzzi, Kleijn, and Zhang}]{DG:Gunbon0}
Ghifary, M.; Balduzzi, D.; Kleijn, W.~B.; and Zhang, M. 2015.
\newblock Domain generalization for object recognition with multi-task autoencoders.
\newblock In \emph{Proceedings of the IEEE international conference on computer vision}, 2551--2559.

\bibitem[{Jian and Torresani(2021)}]{scarcitylabel}
Jian, Y.; and Torresani, L. 2021.
\newblock Label Hallucination for Few-Shot Classification.
\newblock arXiv:2112.03340.

\bibitem[{Khrulkov et~al.(2019)Khrulkov, Mirvakhabova, Ustinova, Oseledets, and Lempitsky}]{episodic:MBML:Hyperbolic}
Khrulkov, V.; Mirvakhabova, L.; Ustinova, E.; Oseledets, I.~V.; and Lempitsky, V.~S. 2019.
\newblock Hyperbolic Image Embeddings.
\newblock \emph{CoRR}, abs/1904.02239.

\bibitem[{Kirkpatrick et~al.(2016)Kirkpatrick, Pascanu, Rabinowitz, Veness, Desjardins, Rusu, Milan, Quan, Ramalho, Grabska{-}Barwinska, Hassabis, Clopath, Kumaran, and Hadsell}]{continual:catastrophic}
Kirkpatrick, J.; Pascanu, R.; Rabinowitz, N.~C.; Veness, J.; Desjardins, G.; Rusu, A.~A.; Milan, K.; Quan, J.; Ramalho, T.; Grabska{-}Barwinska, A.; Hassabis, D.; Clopath, C.; Kumaran, D.; and Hadsell, R. 2016.
\newblock Overcoming catastrophic forgetting in neural networks.
\newblock \emph{CoRR}, abs/1612.00796.

\bibitem[{Krause et~al.(2013)Krause, Stark, Deng, and Fei-Fei}]{dataset:cars}
Krause, J.; Stark, M.; Deng, J.; and Fei-Fei, L. 2013.
\newblock 3d object representations for fine-grained categorization.
\newblock \emph{In Proceedings of the IEEE international conference on computer vision workshops}.

\bibitem[{Lee, Yoo, and Kwak(2023)}]{lee2023shot}
Lee, J.; Yoo, J.; and Kwak, N. 2023.
\newblock {SHOT}: Suppressing the Hessian along the Optimization Trajectory for Gradient-Based Meta-Learning.
\newblock In \emph{Thirty-seventh Conference on Neural Information Processing Systems}.

\bibitem[{Li et~al.(2018)Li, Yang, Yang, and Hospedales}]{DG:Gunbon1}
Li, D.; Yang, Y.; Yang, Y.; and Hospedales, T.~M. 2018.
\newblock Learning to generalize: Meta-learning for domain generalization.
\newblock In \emph{Proceedings of the Thirty-Second AAAI Conference on Artificial Intelligence}.

\bibitem[{Lopez{-}Paz and Ranzato(2017)}]{continual:memoryrehearsal}
Lopez{-}Paz, D.; and Ranzato, M. 2017.
\newblock Gradient Episodic Memory for Continuum Learning.
\newblock \emph{CoRR}, abs/1706.08840.

\bibitem[{Maurer, Pontil, and Romera-Paredes(2016)}]{DG:wisdom2}
Maurer, A.; Pontil, M.; and Romera-Paredes, B. 2016.
\newblock The Benefit of Multitask Representation Learning.
\newblock arXiv:1505.06279.

\bibitem[{Nichol, Achiam, and Schulman(2018)}]{episodic:GBML:Reptile}
Nichol, A.; Achiam, J.; and Schulman, J. 2018.
\newblock On first-order meta-learning algorithms.
\newblock \emph{arXiv preprint arXiv:1803.02999}.

\bibitem[{Oh et~al.(2020)Oh, Yoo, Kim, and Yun}]{episodic:GBML:BOIL}
Oh, J.; Yoo, H.; Kim, C.; and Yun, S.-Y. 2020.
\newblock Boil: Towards representation change for few-shot learning.
\newblock \emph{arXiv preprint arXiv:2008.08882}.

\bibitem[{Raghu et~al.(2019)Raghu, Raghu, Bengio, and Vinyals}]{episodic:GBML:ANIL}
Raghu, A.; Raghu, M.; Bengio, S.; and Vinyals, O. 2019.
\newblock Rapid Learning or Feature Reuse? Towards Understanding the Effectiveness of {MAML}.
\newblock \emph{CoRR}, abs/1909.09157.

\bibitem[{Rajeswaran et~al.(2019)Rajeswaran, Finn, Kakade, and Levine}]{episodic:GBML:imaml}
Rajeswaran, A.; Finn, C.; Kakade, S.~M.; and Levine, S. 2019.
\newblock Meta-learning with implicit gradients.
\newblock \emph{Advances in neural information processing systems}, 32.

\bibitem[{Ren et~al.(2018)Ren, Triantafillou, Ravi, Snell, Swersky, Tenenbaum, Larochelle, and Zemel}]{dataset:Tieredimagenet}
Ren, M.; Triantafillou, E.; Ravi, S.; Snell, J.; Swersky, K.; Tenenbaum, J.~B.; Larochelle, H.; and Zemel, R.~S. 2018.
\newblock Meta-learning for semi-supervised few-shot classification.
\newblock \emph{In the Sixth International Conference on Learning Representations}.

\bibitem[{ROBINS(1995)}]{catastrophygunbon}
ROBINS, A. 1995.
\newblock Catastrophic Forgetting, Rehearsal and Pseudorehearsal.
\newblock \emph{Connection Science}, 7(2): 123--146.

\bibitem[{Russakovsky et~al.(2015)Russakovsky, Deng, Su, Krause, Satheesh, Ma, Huang, Karpathy, Khosla, Bernstein et~al.}]{dataset:imagenet}
Russakovsky, O.; Deng, J.; Su, H.; Krause, J.; Satheesh, S.; Ma, S.; Huang, Z.; Karpathy, A.; Khosla, A.; Bernstein, M.; et~al. 2015.
\newblock Imagenet large scale visual recognition challenge.
\newblock \emph{International journal of computer vision}, 115: 211--252.

\bibitem[{Rusu et~al.(2018)Rusu, Rao, Sygnowski, Vinyals, Pascanu, Osindero, and Hadsell}]{episodic:GBML:LEO}
Rusu, A.~A.; Rao, D.; Sygnowski, J.; Vinyals, O.; Pascanu, R.; Osindero, S.; and Hadsell, R. 2018.
\newblock Meta-Learning with Latent Embedding Optimization.
\newblock \emph{CoRR}, abs/1807.05960.

\bibitem[{Snell, Swersky, and Zemel(2017)}]{episodic:MBML:Protonet}
Snell, J.; Swersky, K.; and Zemel, R.~S. 2017.
\newblock Prototypical Networks for Few-shot Learning.
\newblock \emph{CoRR}, abs/1703.05175.

\bibitem[{Thrun and Pratt(2012)}]{labelequiv:metadefinition}
Thrun, S.; and Pratt, L. 2012.
\newblock \emph{Learning to learn}.
\newblock Springer Science \& Business Media.

\bibitem[{Torralba and Efros(2011)}]{theory:specificdataset}
Torralba, A.; and Efros, A.~A. 2011.
\newblock Unbiased look at dataset bias.
\newblock In \emph{CVPR 2011}, 1521--1528.

\bibitem[{Vinyals et~al.(2016)Vinyals, Blundell, Lillicrap, and Wierstra}]{dataset:MiniImageNet}
Vinyals, O.; Blundell, C.; Lillicrap, T.; and Wierstra, D. 2016.
\newblock Matching networks for one shot learning.
\newblock \emph{Advances in neural information processing systems}, 29.

\bibitem[{Welinder et~al.(2010)Welinder, Branson, Mita, Wah, Schroff, Belongie, and Perona}]{dataset:cub}
Welinder, P.; Branson, S.; Mita, T.; Wah, C.; Schroff, F.; Belongie, S.; and Perona, P. 2010.
\newblock Caltech-UCSD Birds 200.
\newblock \emph{Technical Report CNS-TR-2010-001, California Institute of Technology}.

\end{thebibliography}

\newpage
\appendix
\section{good}

\end{document}